\documentclass[11pt]{article} % For LaTeX2e
\usepackage[utf8]{inputenc} % allow utf-8 input
\usepackage[T1]{fontenc}    % use 8-bit T1 fonts
\usepackage{hyperref}       % hyperlinks
\usepackage{url}            % simple URL typesetting
\usepackage{booktabs}       % professional-quality tables
\usepackage{amsfonts}       % blackboard math symbols
\usepackage{nicefrac}       % compact symbols for 1/2, etc.
\usepackage{microtype}      % microtypography
\usepackage{xcolor}         % colors

\usepackage[linesnumbered,ruled]{algorithm2e}
\usepackage{subcaption}
\usepackage{graphicx}
\usepackage{caption}
\usepackage{float}

\usepackage{mathptmx}      % use Times fonts if available on your TeX system
\usepackage{times}
\usepackage{mathtools}
\usepackage{amsmath,amssymb}
\usepackage{bbm}
\usepackage{dsfont}
\usepackage{comment}
\usepackage[title]{appendix}
\usepackage[english]{babel}
\usepackage[colorinlistoftodos]{todonotes}

\newtheorem{definition}{Definition}[section]

\newtheorem{theorem}{Theorem}[section]
\usepackage{algpseudocode}
\usepackage[export]{adjustbox}

\newcommand{\Rb}{\mathbbm{R}}      % for Real numbers

\newcommand{\Bc}{\mathcal{B}}

\newcommand{\Qc}{\mathcal{Q}}

\newcommand{\Eb}{\mathbbm{E}}

\newcommand{\Lc}{\mathcal{L}}

\newcommand{\Pc}{\mathcal{P}}

\newcommand{\Vc}{\mathcal{V}}
\newcommand{\Xc}{\mathcal{X}}
\newcommand{\Yc}{\mathcal{Y}}
\newcommand{\Ic}{\mathcal{I}}
\newcommand{\Wc}{\mathcal{W}}
\newcommand{\Zc}{\mathcal{Z}}

\newcommand{\argmin}{\mathop{\rm argmin}}

\newcommand{\supp}{\mathop{\rm supp}}

\newcommand{\D}{\mathop{\text{\rm d}\!}}

\renewcommand{\eqref}[1]{\textup{(\ref{#1})}}
\def\1{\mathbbm{1}}

\newenvironment{tightlist}[1]{%
    \list{{\textup{(\Alph{enumi})}}}{\settowidth\labelwidth{{\textup{(#1)}}}
    \leftmargin 12pt \advance\leftmargin\labelsep \itemindent \parindent
    \parsep 0pt plus 1pt minus 1pt \topsep 0pt \itemsep 0pt
    \usecounter{enumi}}}{\endlist}

\title{Fast Dual Subgradient Optimization of the Integrated Transportation Distance Between Stochastic Kernels}

\author{Zhengqi Lin and Andrzej Ruszczy\'nski\footnote{Department of Management Science and Information Systems,
Rutgers University,
Piscataway, NJ o8854, email:
\texttt{zl458@rutgers.edu; rusz@rutgers.edu}}
}

\date{December 1, 2023}

\begin{document}

\maketitle

\begin{abstract}

A generalization of the Wasserstein metric, the integrated transportation distance,  establishes a novel distance between probability kernels of Markov systems. This metric serves as the foundation for an efficient approximation technique, enabling the replacement of the original system's kernel with a kernel with a discrete support of limited cardinality. To facilitate practical implementation, we present a specialized dual algorithm capable of constructing these approximate kernels quickly and efficiently, without requiring computationally expensive matrix operations. Finally, we demonstrate the efficacy of our method through several illustrative examples, showcasing its utility in practical scenarios. This advancement offers new possibilities for the streamlined analysis and manipulation of stochastic systems represented by kernels.

\end{abstract}

\section{Introduction}

An extension of the Wasserstein metric, named the \emph{integrated transportation distance}, was introduced in \cite{lin2023integrated} to establish a novel distance between stochastic (Markov) kernels. The paper considers a discrete-time Markov system described by the relations:
\begin{equation}
\label{kernel-model}
X_{t+1} \sim Q_t(X_t), \quad t=0,1,\dots,T-1,
\end{equation}
where $X_t\in \Xc$ is the state at time $t$, and $Q_t:\Xc \to \Pc(\Xc)$, $t=0,1,\dots,T-1$, are stochastic kernels. The symbol $\Xc$ represents a separable metric space
(the state space), and $\Pc(\Xc)$ is the space of probability measures on $\Xc$.
Formula \eqref{kernel-model} means that the conditional distribution
of $X_{t+1}$, given $X_t=x$, is $Q_t(x)$. The distribution of the initial state $\delta_{x_0}$ (the Dirac delta at $x_0$) and the sequence of kernels $Q_t$, $t=0,\dots,T-1$, define a probability measure $P$ on the  space of paths $\Xc^{T+1}$.

One of the challenges of dealing with models of the form \eqref{kernel-model} is the need to evaluate a backward system (with a sequence of functions $c_t:\Xc\to \Rb$):
\begin{equation}
\label{DP-risk-finite}
\begin{aligned}
v_t(x) = c_t(x) +  \sigma_t\big(x,Q_t(x), v_{t+1}(\cdot)\big),\quad & x\in \Xc, \quad t=0, \dots, T-1;\\
v_T(x) = c_T(x), \quad & x \in \Xc.
\end{aligned}
\end{equation}

In equation \eqref{DP-risk-finite}, the operator $\sigma_t:\Xc \times  \Pc(\Xc)\times \Vc\to \Rb$, where $\Vc$ is a space of Borel measurable real functions on $\Xc$, is a \emph{transition risk mapping}.  Its first argument is the present state~$x$. The second argument is the probability distribution $Q_t(x)$ of the state following $x$ in the system \eqref{kernel-model}. The last argument, the function $v_{t+1}(\cdot)$, is the next state's value: the risk of running the system from the next state in the time interval from $t+1$ to $T$.

A simple case of the transition risk mapping is the bilinear form,
\begin{equation}
\label{sigma-E}
\sigma_t\big(x,\mu, v_{t+1}(\cdot)\big) = \Eb_{\mu} \big[ v_{t+1}(\cdot)\big].
\end{equation}
In this case, the
scheme \eqref{DP-risk-finite} evaluates the conditional expectation of the total cost from stage $t$ to the end of the horizon $T$:
\[
v_t(x) = \Eb\big[ c_t(X_t) + \dots + c_T(X_T)\,\big|\, X_t = x \big], \quad x \in \Xc,\quad t=0,\dots,T.
\]

Problems of this type arise in manifold applications, such as financial option pricing, risk evaluation, and other dynamic programming problems. One of the challenges associated with the backward system \eqref{DP-risk-finite} is the numerical solution in the case when the transition risk mappings $\sigma_t(\cdot,\cdot,\cdot)$  are nonlinear with respect to the probability measures involved. Therefore, \cite{lin2023integrated} illustrates how the integrated transportation distance can serve as the foundation for an efficient approximation technique, allowing the replacement of the original system's kernel with a discrete support of limited cardinality.

Our objective  is to present a computational method based on approximating the kernels $Q_t(\cdot)$ by simpler, easier-to-handle kernels $\widetilde{Q}_t(\cdot)$, and using them in the backward system \eqref{DP-risk-finite}. In \cite{lin2023integrated}, a mixed-integer programming formulation is employed to approximate the kernels. However, for large-scale problems with numerous variables and constraints, integer or even linear programming can become computationally intractable. To address this challenge, our paper introduces a specialized dual algorithm designed for practical implementation. This algorithm efficiently constructs approximate kernels without relying on computationally expensive matrix operations. We demonstrate the effectiveness of our method by showcasing its utility in diverse practical scenarios and comparing its performance with an integer programming solver. This advancement opens up new possibilities for the streamlined analysis and manipulation of Markov systems represented by kernels.

The approximation of stochastic processes in discrete time has attracted the attention of researchers for many decades.
Fundamental in this respect is the concept of a \emph{scenario tree}.  Ref. \cite{hoyland2001generating} uses
statistical parameters, such as moments and correlations, to construct such a tree. Ref. \cite{kaut2011shape}  involves copulas to capture the shape of the
distributions. Ref. \cite{heitsch2009scenario} was probably the first to use probability metrics for reducing large scenario trees. %A concept of a distance between stochastic processes was proposed by \cite{pflug2001scenario}, and used by  \cite{kovacevic2015tree}  to generate scenario trees.
Ref. \cite{pflug2010version} introduced the concept of nested distance, using an extension of the Wasserstein metric for processes; see also  \cite{pflug2015dynamic}. %\cite{pflug2012distance,pflug2015dynamic}.
All these approaches differ from our construction in the Markovian case.

{The Wasserstein distance has shown promising results in various applications such as Generative Adversarial Networks (GAN) \cite{arjovsky2017wasserstein}, clustering \cite{ho2017multilevel}, semi-supervised learning \cite{pmlr-v32-solomon14}, and image retrievals \cite{Rubner2000TheEM,5459199}, among others.}  Some recent contributions measure the distance of mixture distributions rather than kernels.  \cite{SW} propose the sketched Wasserstein distance, a type of distance metric dedicated to finite mixture models. Research on Wasserstein-based distances tailored to Gaussian mixture models is reported in \cite{GMM1, GMM2, GMM3}.

In parallel, we see continuous efforts to develop fast algorithms for computing the relevant transportation distances. One notable contribution is the Sinkhorn algorithm, introduced by \cite{sinkhorn1}, which incorporates an entropic regularization term to the mass transportation problem. Since then, both the Sinkhorn algorithm and its variant Greenkhorn \cite{Greenkhorn} have become the baseline approaches for computing transportation distance and have triggered significant progress \cite{sinkhorn2,APDAMD1}. Other relevant approaches include accelerated primal-dual gradient descent (APDAGD) \cite{APDAGD1,APDAGD2,APDAGD3} and semi-dual gradient descent \cite{semi_dual1,semi_dual2}.

\paragraph{Organization.}
In section \ref{s2}, we provide a brief overview of the distance metrics, including the Wasserstein and the integrated transportation distances. We also introduce the problem of selecting representative particles using a mixed-integer formulation based on distance metrics. In section \ref{s3}, we present our subgradient method, its relation to the dual problem, and the algorithm used for selecting particles. In section \ref{s4}, we provide a numerical example featuring a 2-dimensional and 1-time-stage Gaussian distribution. We chose this straightforward case to enhance the visualization of outcomes, facilitating effective method comparisons, and bringing attention to the limitations of Mixed-Integer Programming (MIP) solvers in more scenarios.
Section \ref{conclusion}  concludes the paper.

\section{The problem}
\label{s2}

\subsection{Wasserstein distance}

Let  $d(\cdot,\cdot)$ be the metric on $\Xc$. For two probability measures $\mu, \nu$ on $\Xc$ having finite moments up to order $p \in[1, \infty)$, their Wasserstein distance of order $p$ is defined by the following formula
(see \cite{rachev1998mass,villani2009optimal} for a detailed exposition and historical account):
\begin{equation}
\label{Wasser}
W_{p}(\mu, \nu) =\left(\inf _{\pi \in \Pi(\mu, \nu)} \int_{\mathcal{X} \times \mathcal{X} } d(x, y)^{p} \;  \pi({\D x}, {\D y})\right)^{1 / p},
\end{equation}
where $\Pi(\mu, \nu)$ is the set of all probability measures in $\Pc(\Xc\times\Xc)$ with the marginals $\mu$ and $\nu$.
%The measure $\pi^* \in \Pi(\mu, \nu)$ that realizes the infimum in Eq. \eqref{Wass} is called the \emph{optimal coupling} or the \emph{optimal transport plan}. It always exists.

%On the space $\Pc(\Xc)$, the functional $W_{p}(\cdot,\cdot)$ is not a metric in the strict sense, because it might take the value $+\infty$; but it does satisfy the axioms of a distance.
We restrict the space of probability measures to measures with finite moments
up to order $p$. Formally, we define the Wasserstein space:
\[
\Pc_{p}(\mathcal{X}):=\left\{\mu \in \Pc(\mathcal{X}) : \ \int_{\mathcal{X}} d\left(x_{0}, x\right)^{p} \;\mu(d x)<+\infty\right\}.
\]
%The Wasserstein space $\Pc\emph{}_{p}(\mathcal{X})$ is therefore the space of probability measures that have a finite moment of order $p$.
For each $p\in [1,\infty)$, the function $W_{p}(\cdot,\cdot)$ defines a metric on $\Pc_{p}(\mathcal{X})$. Furthermore, for all $\mu,\nu\in \Pc_{p}(\mathcal{X})$ the optimal coupling
realizing the infimum in \eqref{Wasser} exists. From now on, $\Pc_{p}(\mathcal{X})$ will be always equipped with the distance $W_p(\cdot,\cdot)$.

For discrete measures, problem \eqref{Wasser} has a linear programming representation.
Let $\mu$ and $\nu$  be supported at positions $\{x^{(i)}\}_{i=1}^{N}$ and $\{z^{(k)}\}_{k=1}^{M}$, respectively, with normalized (totaling 1) positive weight vectors $w_{x}$ and $w_{z}$:
$\mu=\sum_{i=1}^{N} w_{x}^{(i)} \delta_{x^{(i)}}$, $\nu=\sum_{k=1}^{M} w_{z}^{(k)} \delta_{z^{(k)}}$.
For $p \geq 1$, let $D \in {R}_{+}^{N \times M}$ be the distance matrix with elements $d_{ik}=d\big(x^{(i)},z^{(k)}\big)^{p}$. Then the $p$th power of the $p$-Wasserstein distance between the measures $\mu$ and $\nu$ is the optimal value of the following transportation problem:
\begin{equation}
\label{LP-Wass}
\min _{\pi \in {R}_{+}^{N \times M}} \textstyle{\sum_{i=1}^N\sum_{k=1}^M}   d_{i k} \pi_{i k}\quad
\text { s.t.} \quad  \pi^\top \1_{N}=w_{x}, \quad  \pi \1_{M}=w_{z}.
\end{equation}
The calculation of the distance is easy when the linear programming problem \eqref{LP-Wass} can be solved. For large instances, specialized algorithms such as
\cite{sinkhorn1,sinkhorn2,Greenkhorn,APDAMD1,APDAGD1,APDAGD2,APDAGD3,semi_dual1,semi_dual2}
have been proposed. Our problem, in this special case, is more complex: \emph{find $\nu$ supported on a set of the cardinality $M$ such that $W_{p}(\mu, \nu)$ is the smallest possible.} We elaborate on it in the next section.

\subsection{The Integrated Transportation Distance Between Kernels}
\label{s:ITD}

%We further generalize the transportation (Wasserstein) metric between probability distributions in the space of kernels.   By the measure disintegration formula, every probability measure $\mu\in \Pc(\Xc\times\Yc)$ admits a disintegration
%$\mu = {\lambda} \circledast Q$, where ${\lambda} \in \Pc(\Xc)$ is the marginal distribution on $\Xc$, and $Q:\Xc \to \Pc(\Yc)$ is a \emph{kernel} (a function
%such that for each $B\in \Bc(\Yc)$ the mapping $x\mapsto Q(B|x)$ is Borel measurable):
%\[
%\mu(A \times B) = \int_A  Q(B|x)\;{\lambda}({\D x}), \quad \forall \big(A\in \Bc(\Xc)\big),\;\forall \big(B \in \Bc(\Yc)\big).
%\]

Suppose $\Xc$ and $\Yc$ are Polish spaces.  By the measure disintegration formula, every probability measure $\mu\in \Pc(\Xc\times\Yc)$ admits a disintegration
$\mu = {\lambda} \circledast Q$, where ${\lambda} \in \Pc(\Xc)$ is the marginal distribution on $\Xc$, and $Q:\Xc \to \Pc(\Yc)$ is a \emph{kernel} (a function
such that for each $B\in \Bc(\Yc)$ the mapping $x\mapsto Q(B|x)$ is Borel measurable):
\[
\mu(A \times B) = \int_A  Q(B|x)\;{\lambda}({\D x}), \quad \forall \big(A\in \Bc(\Xc)\big),\;\forall \big(B \in \Bc(\Yc)\big).
\]
Conversely, given a marginal ${\lambda} \in \Pc(\Xc)$  and a kernel $Q:\Xc \to \Pc(\Yc)$, the above formula defines a probability measure ${\lambda} \circledast Q$
on $\Xc \times \Yc$.
Its marginal on $\Yc$ is the \emph{mixture distribution} $\lambda \circ Q$ given by
\[
(\lambda \circ Q)(B) = \int_\Xc  Q(B|x)\;{\lambda}({\D x}), \quad \forall \, B \in \Bc(\Yc).
\]

To define a metric between kernels, we restrict the class of kernels under consideration to
the set $\Qc_p(\Xc)$ of kernels $Q:\Xc \to \Pc_p(\Xc)$ such that for each a constant $C$ exists, with which
\[
 \int_\Xc d(y,y_0)^p\; Q({\D y}|x) \le C\big(1 + d(x,x_0)^p\big), \quad \forall \,x\in \Xc.
\]
The choice of the points $x_0\in \Xc$ and $y_0\in \Yc$ is irrelevant, because $C$ may be adjusted.

\begin{definition}
\label{d:kernel-distance}
The integrated transportation distance of order $p$ between two kernels $Q$ and $\widetilde{Q}$ in $\Qc_p(\Xc)$ with a fixed marginal ${\lambda}\in \Pc_p(\Xc)$  is defined as
\[
    \Wc_{p}^\lambda(Q, \widetilde{Q})=\left(\int_{\Xc} \big[{W}_{p}(Q(\cdot | x), \widetilde{Q}(\cdot | x))\big]^p \;{\lambda}(\D x)\right)^{1/p} .
\]
\end{definition}

For a fixed marginal $\lambda\in \Pc_p(\Xc)$, we identify the kernels $Q$ and $\widetilde{Q}$
if ${W}_{p}(Q(\cdot | x), \widetilde{Q}(\cdot | x))=0$ for $\lambda$-almost all $x\in \Xc$. In this way, we define the space $\Qc_p^\lambda(\Xc,\Yc)$ of equivalence classes of $\Qc_p(\Xc,\Yc)$.

\begin{theorem} \label{TS_T}
For any $p\in [1,\infty)$ and any $\lambda\in \Pc_p(\Xc)$, the function $\Wc_p^\lambda(\cdot,\cdot)$, defines a  metric on the space $\Qc_p^\lambda(\Xc,\Yc)$.
\end{theorem}

For a kernel $Q \in \Qc_p(\Xc,\Yc)$, and every $\lambda \in \Pc_p(\Xc)$ the measure $\lambda \circ Q$ is an element of $\Pc_p(\Yc)$, because
\begin{multline*}
\int_\Yc d(y,y_0)^p\; (\lambda\circ Q)({\D y}) = \int_\Xc \int_\Yc d(y,y_0)^p\;\; Q({\D y}|x)\;\lambda({\D x})
 \le C(Q) \int_\Xc  \big(1 + d(x,x_0)^p\big)\;\lambda({\D x}) < \infty.
\end{multline*}
In a similar way, the measure ${\lambda} \circledast Q \in \Pc_p(\Xc\times\Yc)$, because
\begin{multline*}
\int_\Xc\int_\Yc \big[d(x,x_0)^p+d(y,y_0)^p\big]\;  Q({\D y}|x)\;\lambda({\D x})
= \int_\Xc \bigg[d(x,x_0)^p+ \int_\Yc d(y,y_0)^p\;\; Q({\D y}|x)\bigg]\;\lambda({\D x}) \\
 \le (C(Q)+1) \int_\Xc  \big(1 + d(x,x_0)^p\big)\;\lambda({\D x}) < \infty.
\end{multline*}

The integrated transportation distance provides an upper bound on the distances between two mixture distributions and between two composition distributions.

\begin{theorem} \label{TS_T2}
For all $\lambda \in \Pc_p(\Xc)$ and all $Q ,  \widetilde{Q} \in \Qc_p^\lambda(\Xc,\Yc)$,
\[
\Wc_p^\lambda(Q,\widetilde{Q}) \geq {W}_p({\lambda} \circledast Q,{\lambda} \circledast \widetilde{Q}) \geq
{W}_p({\lambda} \circ Q,{\lambda} \circ \widetilde{Q}).
\]
\end{theorem}

The integrated transportation distance can be used to approximate the system \eqref{kernel-model} by a system with finitely supported kernels.
Suppose at stage $t$
%we already have finite sets of \emph{representative points} $\Xc_\tau = \big\{ z_\tau^s\big\}_{s=1,\dots,M_\tau}$, which serve as  discrete representations of the state space at $\tau$.
we already have for all $\tau=0,\dots,t-1$  approximate kernels $\widetilde{Q}_\tau:\Xc \to\Pc(\Xc)$. These kernels define the approximate marginal distribution
\[
\widetilde{\lambda}_t = \delta_{x_0}\circ \widetilde{Q}_0 \circ \widetilde{Q}_1 \circ \dots \circ \widetilde{Q}_{t-1} = \widetilde{\lambda}_{t-1}\circ \widetilde{Q}_{t-1}.
\]
We also have the finite subsets  $\Xc_\tau = \text{supp}(\widetilde{\lambda}_\tau)$, $\tau=0,1,\dots,t$.
For $t=0$, $\widetilde{\lambda}_0=\delta_{x_0}$, and $\Xc_0=\{x_0\}$.

At the stage $t$, we construct a kernel $\widetilde{Q}_t:\Xc_t \to \Pc_p(\Xc)$ such that
\begin{equation}
\label{Delta}
\Wc_p^{\widetilde{\lambda}_t}(Q_t,\widetilde{Q}_t)\le \varDelta_t.
\end{equation}
If $t< T-1$, we increase $t$ by one, and continue; otherwise, we stop.
Observe that the approximate marginal distribution $\widetilde{\lambda}_t$ is well-defined at each step of this abstract scheme.

We then solve the approximate version of the risk evaluation algorithm \eqref{DP-risk-finite}, with the true kernels $Q_t$
replaced by the approximate kernels $\widetilde{Q}_t$, $t=0,\dots,T-1$:
\begin{equation}
\label{DP-risk-approximate}
\widetilde{v}_t(x) = c_t(x) +  \sigma_t\big(x,\widetilde{Q}_t(x), \widetilde{v}_{t+1}(\cdot)\big),\quad x\in \Xc_t, \quad t=0,1,\dots,T-1;
\end{equation}
we assume that $\widetilde{v}_T(\cdot)\equiv v_T(\cdot)\equiv c_T(\cdot)$.

To estimate the error of this evaluation in terms of the kernel errors $\varDelta_t$,
 we make the following general assumptions.

\begin{tightlist}{ii}
\item For every $t=0,1,\dots,T-1$ and for every $x\in \Xc_t$, the operator $\sigma_t(x,\,\cdot\, , v_{t+1})$ is Lipschitz continuous
with respect to the metric $W_p(\cdot,\cdot)$ with the constant $L_t$:
\[
 \big| \sigma_{t}\big(x,\mu, {v}_{t+1}(\cdot)\big) - \sigma_{t}\big(x,\nu, {v}_{t+1}(\cdot)\big)\big|\\
\le L_{t}\, W_p(\mu,\nu), \quad \forall\, \mu,\nu \in \Pc_p(\Xc);
\]
\item For every $x\in \Xc_t$ and for every $t=0,1,\dots,T-1$, the operator $\sigma_t(x,\widetilde{Q}_t(x), \,\cdot\,)$ is
Lipschitz continuous
with respect to the norm in the space $\Lc_p(\Xc,\Bc(\Xc),\widetilde{Q}_t(x))$ with the constant $K_t$:
\[
 \big| \sigma_{t}\big(x,\widetilde{Q}_t(x), {v}(\cdot)\big) - \sigma_{t}\big(x,\widetilde{Q}_t(x), {w}(\cdot)\big)\big|
\le K_{t}\, \|v - w\|_p,\quad  \forall\,v,w\in \Lc_p(\Xc,\Bc(\Xc),\widetilde{Q}_t(x)).
\]
\end{tightlist}
\begin{theorem}
\label{t:error_estimate}
If the assumptions \textup{(A)} and \textup{(B)} are satisfied, then for all  $t=0,\dots,T-1$ we have
\begin{equation}
\label{error-p}
\bigg(\int_\Xc \big| \widetilde{v}_{t}(x) - v_{t}(x) \big|^p\;\widetilde{\lambda}_t({\D x})\bigg)^{1/p} \le \sum_{\tau=t}^{T-1}L_\tau\bigg(\prod_{j=t}^{\tau-1}K_j\bigg)
\varDelta_\tau.
%\Wc_p^{\widetilde{\lambda}_{\tau}}(\widetilde{Q}_{\tau},{Q}_{\tau}).
\end{equation}
\end{theorem}

In order to accomplish \eqref{Delta}, at stage $t$, we  construct a finite set $\Xc_{t+1} \subset \Xc$ of  cardinality $M_{t+1}$ and a kernel $\widetilde{Q}_t:\Xc_t \to \Pc(\Xc_{t+1})$
by solving the following  problem:
\begin{equation}
\label{kernel-search}
\begin{aligned}
\min_{\Xc_{t+1},\widetilde{Q}_t} &\  \Wc_p^{\widetilde{\lambda}_t}(Q_t,\widetilde{Q}_t)\\
\text{s.t.} &\ \supp(\widetilde{\lambda}_t \circ \widetilde{Q}_t) = \Xc_{t+1},\\
&\ \big| \Xc_{t+1} \big| \le M_{t+1}.
\end{aligned}
\end{equation}
The cardinality $M_{t+1}$ has to be chosen experimentally, to achieve the desired accuracy in \eqref{Delta}.
After (approximately) solving this problem, we increase $t$ by one and continue.

Let us focus on effective ways for constructing an approximate solution to problem \eqref{kernel-search}.
We represent the (unknown) support of $\widetilde{\lambda}_t \circ \widetilde{Q}_t$ by $\Xc_{t+1} = \big\{ z_{t+1}^\ell\big\}_{\ell=1,\dots,M_{t+1}}$
and the (unknown) transition probabilities by $\widetilde{Q}_t(z_{t+1}^\ell|z_t^s)$, $s=1,\dots,M_n$, $\ell=1,\dots,M_{n+1}$.
With the use of the kernel distance,   problem \eqref{kernel-search} can be equivalently rewritten as:
\begin{equation}
\label{kernel-search-2}
\begin{aligned}
\min_{\Xc_{t+1},\widetilde{Q}_t} & \; \sum_{s=1}^{M_n} \widetilde{\lambda}_t^s W_p\big(Q_t(\cdot|z_t^s),\widetilde{Q}_t(\cdot|z_t^s)\big)^p\\
\text{s.t.}&\; \supp \big(\widetilde{Q}_t(\cdot|z_t^s)\big) \subset \Xc_{t+1},\quad s=1,\dots,M_n,\\
&\ \big| \Xc_{t+1} \big| \le M_{t+1}.
\end{aligned}
\end{equation}
 In our  approach, we represent each distribution $Q_t(\cdot|z_t^s)$  by a finite number of particles  $\big\{x_{t+1}^{s,i}\big\}_{i\in \Ic_{t+1}^s}$ drawn independently from $Q_t(\cdot|z_t^s)$. The expected error of this approximation is well-investigated by \cite{dereich2013constructive} and \cite{fournier2015rate} in terms of the sample size $\big| \Ic_{t+1}^s\big|$, the state space dimension, and the distribution's moments. Assuming the error of this large-size discrete approximation as fixed, we aim to construct a smaller support with as little error as possible to the particle distribution. For this purpose, we introduce the sets
$\Zc_{t+1} =\big\{\zeta^k_{t+1}\big\}_{k=1,\dots,K_{t+1}}$.
Each consists of pre-selected potential locations for the next-stage representative states $z_{t+1}^j$, where $j=1,\dots,M_{t+1}$. It may be the union of the sets of particles, $\big\{ x_{t+1}^{s,i}, \ i\in \Ic_{t+1}^s,\ s=1,\dots,M_t\big\}$; often, computational expediency requires that $K_{t+1} < \sum_{s=1}^{M_t} \big|\Ic_{t+1}^s\big|$, we still have $M_{t+1} \ll K_{t+1}$, which makes the task of finding the best representative points challenging.

If the next-stage representative points $\big\{ z_{t+1}^j\big\}_{j=1,\dots,M_{t+1}}$ were known, the problem would have a straightforward solution. For each particle $x_{t+1}^{s,i}$
we would choose the closest representative point,
\[
j^*(i)= \argmin_{j=1,\dots,M_{t+1}} d\big( x_{t+1}^{s,i} , z_{t+1}^j\big),
\]
and set the transportation probabilities $\pi_t^{s,i,j^*(k)}  = \frac{1}{|\Ic_{t+1}^s|}$;
for other $j$, we set them to 0. The implied approximate kernel is
$\widetilde{Q}_t(z_{t+1}^j|z_t^s) = \sum_{i\in \Ic_{t+1}^s} \pi_t^{s,i,j}$, $s=1,\dots,M_t$, $j=1,\dots,M_{t+1}$;
it is the proportion of the particles from $\Ic_{t+1}^s$ assigned to $z_{t+1}^j$.

To find the best representative points, we introduce the binary variables
\[
\gamma_k = \begin{cases} 1 & \text{ if the point $\zeta_{t+1}^k$ has been selected to $\Xc_{t+1}$},\\
0 & \text{ otherwise},
\end{cases}
\quad k=1,\dots,K_{t+1},
\]
and we re-scale the transportation plans:
\[
\beta_{sik} = |\Ic_{t+1}^s|\pi_t^{s,i,k}, \quad s=1,\dots,M_t,\  i \in \Ic_{t+1}^s,\  k=1,\dots,K_{t+1}.
\]
We obtain from \eqref{kernel-search-2} the following linear mixed-integer optimization problem (we omit the ranges of the sums when they are evident):
\begin{subequations}
\label{mixed-bin}
\begin{align}
\min_{\gamma,\beta}
&\; { \sum_{s}  w_s \sum_{i}\sum_{k} d_{sik} \beta_{sik} }\label{mb-a}\\
\text{s.t.}
&\; \beta_{sik} \in[0,1],\  \gamma_k\in \{0,1\}, \quad s=1,\dots,M_t,\quad  i \in \Ic_{t+1}^s,\quad  k=1,\dots,K_{t+1},\label{mb-e}\\
&\;\beta_{sik} \le \gamma_k,\quad s=1,\dots,M_t,\quad i \in \Ic_{t+1}^s,\quad k=1,\dots,K_{t+1}, \label{mb-b}\\
&\; {\sum_{k} \beta_{sik} = 1, \quad s=1,\dots,M_t,\quad i \in \Ic_{t+1}^s}, \label{mb-c}\\
&\; {\sum_{k} \gamma_k  \le M_{t+1}},\label{mb-d}
\end{align}
\end{subequations}
with $w_s = \frac{\widetilde{\lambda}_t^s}{|\Ic_{t+1}^s|}$ and  $d_{sik}=d\big( x_{t+1}^{s,i} , \zeta_{t+1}^k\big)^p$.
The implied approximate kernel is:
\begin{equation}
\label{implied-kernel-2}
{ \widetilde{Q}_t(z_{t+1}^k|z_t^s) = \frac{1}{|\Ic_{t+1}^s|}\sum_{i} \beta_{sik}},\quad s=1,\dots,M_t,\quad k=1,\dots,M_{t+1}.
\end{equation}
Finally, $\widetilde{\lambda}_{t+1} = \widetilde{\lambda}_t \circ \widetilde{Q}_t$, and the iteration continues until $t=T-1$.
%Later, we simplify $\frac{\widetilde{\lambda}_t^s}{|\Ic_{t+1}^s|}$ as $w_s$.

Since problem \eqref{mixed-bin} involves binary variables, it is reasonable to employ an integer programming solver, such as Gurobi, CPLEX, or SCIP. However, integer or even linear programming can become computationally intractable for large-scale problems with many variables and constraints. Therefore, in section \ref{s3}, we propose a subgradient-based method to solve problem \eqref{mixed-bin}.
%Nonetheless, it is important to note that the subgradient method does not always guarantee convergence to the optimal solution and may require more iterations to achieve a %satisfactory result compared to integer programming.

The particle selection problem using the Wasserstein distance is a simplified form of problem \eqref{mixed-bin}. In the case of $M_t=1$, we obtain the problem of finding the best $\nu$ in \eqref{LP-Wass}. Notably, the facility location and clustering problems share similarities with our particle selection method as well.
\begin{comment}These problems involve finding optimal locations of facilities or grouping objects based on a cost or distance function. The facility location problem seeks to determine the best locations for facilities to serve a given set of demand points while minimizing the total cost of servicing them. The clustering problem involves grouping similar objects together based on a distance metric while minimizing the total cost of forming the clusters. These problems share similarities with particle selection, as they all aim to find optimal locations or groups based on a distance or cost function. Our proposed method can be adapted to these problems by defining a cost or distance function based on the distance between the locations of facilities or objects. We also demonstrate the effectiveness of our method on the "mixture Gaussian distribution" example in section 4.1, where we show significant improvement in computational efficiency compared to an established mixed integer programming solver.
\end{comment}

\section{Dual subgradient method}
\label{s3}

In this section, we propose a subgradient algorithm to address the computational intractability of large-scale instances of problem \eqref{mixed-bin}. While the subgradient method does not ensure convergence to the strictly optimal solution, it is faster than the mixed-integer linear programming approach and it scales better. We use the fact that our primary objective is to determine the $\gamma$'s, which the subgradient method can effectively accomplish. We present the dual problem in Section \ref{s3.1}, and the exact algorithm used for selecting particles in Section  \ref{s3.2}.

\subsection{The dual problem}

\label{s3.1}

Assigning Lagrange multipliers $\theta_{si}$ and $\theta_0 \ge 0$ to the constrains \eqref{mb-c} and \eqref{mb-d}, respectively,
%$\sum_{k=1}^{K_{t+1}} \beta^{s,i,k} = 1$ for $ s=1,\dots,M_t$, $i \in \Ic_{t+1}^s,$ and $\sum_{k=1}^{K_{t+1}} \gamma_k = M_{t+1}$,
we obtain the Lagrangian function of problem \eqref{mixed-bin}:
\[
 L(\gamma,\beta;\theta)
 =  \sum_{s}\sum_{i}\sum_{k} w_s d_{sik} \beta_{sik}
 + \sum_{s}\sum_{i}\theta_{si} \big(1 - \sum_{k} \beta_{sik} \big)   +\theta_{0} \big(\sum_{k} \gamma_k -  M_{t+1}  \big).
\]
The dual variable $\theta_0$ has the interpretation of the marginal contribution of an additional point to reducing the kernel distance. The variables
$\theta_{si}$ serve as thresholds in the assignment of the particles $x_{t+1}^{s,i}$ to the candidate points. They are needed for the algorithm but are not
used in the final assignment, which can be done easily once the $\gamma$'s are known.
The corresponding dual function is
\begin{equation}
\begin{aligned}
L_D(\theta)  & = \min_{\gamma,\beta \in \Gamma} L(\gamma,\beta; \theta) \\
& = \sum_{k=1}^{K_{t+1}} \, \bigg\{ \min_{\gamma_k,\beta_{\cdot\cdot k} \in \Gamma_k} \,  \sum_{s=1}^{M_t}\sum_{i\in \Ic_{t+1}^s}(w_s d_{sik} - \theta_{si})\beta_{sik} + \theta_{0}\gamma_k \, \bigg\}  + \sum_{s=1}^{M_t}\sum_{i\in \Ic_{t+1}^s}\theta_{si} - M_{t+1}\theta_{0},
\end{aligned}
\label{dual}
\end{equation}
where $\Gamma$ is the feasible set of the primal variables given by the conditions \eqref{mb-e}--\eqref{mb-b}, and $\Gamma_k$ is its projection on the
subspace associated with the $k$th candidate point $\zeta_{t+1}^k$.
The minimization in \eqref{dual} decomposes into $K_{t+1}$ subproblems, each having a closed-form solution. We can perform these calculations in parallel, which provides a significant computational advantage and reduces the optimization time. We see that
$\beta_{sik} =  1$, if $\gamma_k = 1$ and $\theta_{si} > w_s d_{sik}$; it may be arbitrary in $[0,1]$, if $\gamma_k = 1$ and exact equality is satisfied; and is 0, otherwise.
Therefore, for all $k=1,\dots,K_{t+1}$,
\[
\gamma_k  =  1, \quad  \text{ if }\quad  \theta_0 < \displaystyle{\sum_{s=1}^{M_t}\sum_{i\in \Ic_{t+1}^s}}\max(0,\theta_{si} -w_s d_{sik});
\]
$\gamma_k \in\{0,  1\}$, if exact equality holds; and $\gamma_k=0$, otherwise. We denote by $\hat{\Gamma}(\theta)$ the set of solutions of
problem \eqref{dual}. It is worth stressing that in the algorithm below, we need only \emph{one} solution for each $\theta$.

The dual problem has the form
\begin{equation}
\label{dual-problem}
\max_{\theta} L_D(\theta), \quad \text{s.t.}\quad \theta_0 \ge 0.
\end{equation}
The optimal value of \eqref{dual-problem} may be strictly below the optimal value of \eqref{mixed-bin};
it is equal to the optimal value of the linear programming relaxation, where the conditions $\gamma_k\in\{0,1\}$ are replaced by $\gamma_k\in [0,1]$. However, if we replace $M_{t+1}$ by the number of $\gamma_k$'s equal to 1, the gap is zero. If we keep $M_{t+1}$ unchanged, we can construct a feasible solution by setting to 0 the $\gamma_k$'s for which the change in the expression in the braces in \eqref{dual} is the smallest. This allows for the estimation of the gap.

The subdifferential of the dual function has the form
\begin{equation}
\label{subdifferential}
\partial L_D(\theta)=\operatorname{conv}\left\{ \begin{bmatrix}
\left\{1-\sum_{k=1}^{K} \hat{\beta}_{sik} \right\}_{s=1,\dots,M_t,\  i \in \Ic_{t+1}^s }\\
\sum_{k=1}^{K} \hat{\gamma}_k - M_{t+1} \end{bmatrix}: (\hat{\gamma},\hat{\beta})\in \hat{\Gamma}(\theta) \right\}.
\end{equation}
At the optimal solution $\hat{\theta}$ we have $0 \in \partial L_D(\hat{\theta})$, because $\hat{\theta_0}>0$ (the constraint \eqref{mb-d} must be active).

\subsection{The algorithm}

\label{s3.2}

In Algorithm \ref{a2}, we use $j$ to denote the iteration number, starting from 0. The variable $\theta$ represents the initial values of the dual variables, while $M$ represents the number of desired grid points. The parameter $\epsilon$ specifies the tolerance level. The value $\alpha^{(0)}$ denotes the initial learning rate. The variables $\varkappa_1$ and $\varkappa_2$ are exponential decay factors between 0 and 1, which determine the relative contribution of the current gradient and earlier gradients to the direction.  It is important to note that the total number of $\gamma$'s selected by the subgradient method may not necessarily equal $M$, when the stopping criteria are met. However, for the particle selection method, the constraint $\sum_{k=1}^{K} \gamma_k \le M_{t+1}$ is not strictly enforced (it is a modeling issue). We end the iteration when $\sum_{k=1}^{K} \gamma_k$ is close to $M_{t+1}$.

For the (approximate) primal recovery, we choose  $J$ last values $\theta^{(j)}$ at which $L_D(\theta^{(j)})$ is near optimal, and consider the convex hull
of the observed subgradients of the dual function at these points as an approximation of the subdifferential \eqref{subdifferential}. The minimum norm element
in this convex hull corresponds to a convex combination of the corresponding dual points:
$(\bar{\gamma},\bar{\beta}) = \sum_{j\in J} \omega_j(\gamma^{(j)},\beta^{(j)})$, with $\sum_{j\in J} \omega_j = 1$, $\omega_j \ge 0$.

By the duality theory in convex optimization, if the subgradients were  collected at the optimal point, $(\bar{\gamma},\bar{\beta})$ would be the solution
of the convex relaxation of \eqref{mixed-bin}. So,
if the norm of the convex combination of the subgradients is small, then $\sum_{k=1}^{K} \bar{\gamma}_k \approx M_{t+1}$, and we may regard $\bar{\gamma}$ as an approximate solution. We interpret it as the best ``mixed strategy'' and select each point $k$ with probability $\bar{\gamma}_k$. In our experiments, we simply use $\omega_j =
\big( \sum_{i\in J}\alpha^{(i)}\big)^{-1}\alpha^{(j)}\approx 1/|J|$.
This approach is well supported theoretically by \cite{larsson1999ergodic}. The $\mathcal{O}(1/\sqrt{j+1})$ rate of convergence of the subgradient method is well understood since \cite{zinkevich2003online} (see also the review by \cite{garrigos2023handbook}).

\begin{algorithm}[h]
    \SetKwInOut{Input}{Input}
    \SetKwInOut{Output}{Output}
    \Input{ $\theta^{(0)}$, $M$, $\epsilon$, $\alpha^{(0)}$,$\varkappa_1$,$\varkappa_2$ and $j=0$.}
    \Output{ $\theta$,$\gamma$, and $\beta$.}
    \While{$ \sum_{k=1}^{K} \gamma_k < (1-a)*M$  \text{\rm \textbf{or}} $\sum_{k=1}^{K}\gamma_k > (1+a)*M$ \text{\rm \textbf{or}} $\| L_D{(\theta^{(j)})} - L_D{(\theta^{(j-1)})} \| > \epsilon  $}{
    \For{$k = 1, \dots, K$}{
    \eIf{$\sum_{s=1}^{N}\sum_{i\in \Ic^s} \max(0,\theta_{si} -w_s d_{sik}) > \theta_0$  }{
    $\gamma_k \leftarrow 1$\;
    $\beta_{sik} \leftarrow \mathbf{1}_{\{w_s d_{sik} < \theta_{si}^{(j)}\}} $,\quad  $s = 1, \dots, N$,\  $i\in \Ic^s$\;
   }
      {
    $\gamma_k \leftarrow 0$\; \par
    $\beta_{sik} \leftarrow 0$, \quad $   s = 1, \dots, N$, $i\in \Ic^s$\;
    }
    }
    $ \alpha^{(j+1)} \leftarrow \frac{\alpha^{(0)}}{\sqrt{j+1}}$; \par
    $m^{(j+1)}_0 \leftarrow (1-\varkappa_1)(\sum_{k=1}^{K} \gamma_k-M) + \varkappa_1m^{(j)}_{0}$\; \par
    $\theta^{(j+1)}_{0} \leftarrow \theta^{(j)}_{0} + \alpha^{(j+1)} m^{(j+1)}_{0}$\; \par
    $m^{(j+1)}_{si} \leftarrow (1-\varkappa_2)(1-\sum_{k=1}^{K} \beta_{sik}) + \varkappa_2 m^{(j)}_{si}$\; \par
    $\theta^{(j+1)}_{si} \leftarrow \theta^{(j)}_{si} + \alpha^{(j+1)} m^{(j+1)}_{si}$ $ \quad  s = 1, \dots, N$, $i\in \Ic^s$\; \par
    $j \leftarrow j + 1\; $
    }
    \caption{Dual subgradient method with momentum}
     \label{a2}
\end{algorithm}
In the stochastic version of the method, the loop over $k$ in lines 2--10 is executed in a randomly selected batch $\Bc^{(j)} \subset\{1,\dots,K\}$
of size $B\ll K$. Then, in line 12, the subgradient component $g_0^{(j)} = \sum_{k=1}^{K} \gamma_k-M$ is replaced by its stochastic estimate
$\tilde{g}_0^{(j)} = (K/B)\sum_{k\in \Bc^{(j)}} \gamma_k-M$. In line 14, the subgradient components
$g_{si}^{(j)} = 1-\sum_{k=1}^{K} \beta_{sik}$ are replaced by their estimates $\tilde{g}_{si}^{(j)} = 1-(K/B)\sum_{k\in \Bc^{(j)}} \beta_{sik}$.
If the batches are independently drawn at each iteration, the algorithm is a version of the stochastic subgradient method with momentum
(see \cite{yan2018unified,liu2020improved} and the references therein).

\section{Numerical illustration}
\label{s4}

%\subsection{Mixture Gaussian distribution}
%\label{s4.1}

\begin{figure}[H]
\centering

\begin{subfigure}{0.3\textwidth}
    \includegraphics[width=\textwidth]{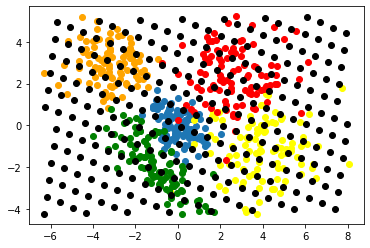}
  \caption{500 sample points and 256 potential representative points}
  \label{fig:p11}
\end{subfigure}
\hfill
\begin{subfigure}{0.3\textwidth}
    \includegraphics[width=\textwidth]{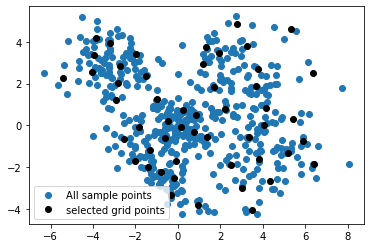}
  \caption{Mixed-Integer Solver (Gurobi with a Python extension) }
  \label{fig:p12}
\end{subfigure}
\hfill
\begin{subfigure}{0.3\textwidth}
    \includegraphics[width=\textwidth]{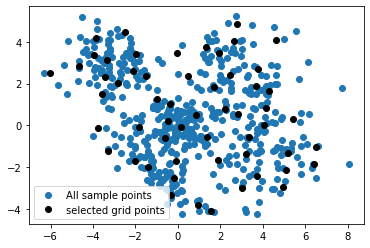}
  \caption{Subgradient method with momentum}
  \label{fig:p13}
\end{subfigure}
\hfill

\caption{$\text{dim}(\beta) = 128000$, $\text{dim}(\gamma) = 256$, and 51 selected particles.}
\label{fig:P1}
\end{figure}

\begin{figure}[H]
\centering
\begin{subfigure}{0.3\textwidth}
    \includegraphics[width=\textwidth]{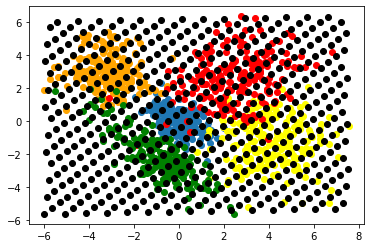}
  \caption{1000 sample points and 512 potential representative points}
  \label{fig:p21}
\end{subfigure}
\hfill
\begin{subfigure}{0.3\textwidth}
    \includegraphics[width=\textwidth]{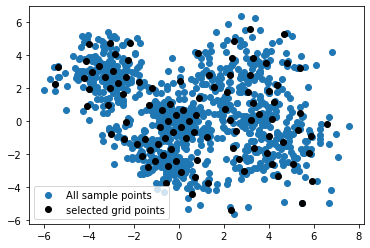}
  \caption{Mixed-Integer Solver (Gurobi with a Python extension) }
  \label{fig:p22}
\end{subfigure}
\hfill
\begin{subfigure}{0.3\textwidth}
    \includegraphics[width=\textwidth]{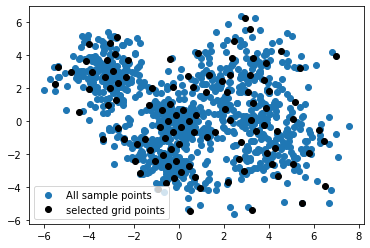}
  \caption{Subgradient method with momentum}
  \label{fig:p23}
\end{subfigure}
\hfill

\caption{$\text{dim}(\beta) = 512000$, $\text{dim}(\gamma) = 512$, and 102 selected particles}
\label{fig:P2}
\end{figure}

\begin{figure}[H]
\centering

\begin{subfigure}{0.3\textwidth}
    \includegraphics[width=\textwidth]{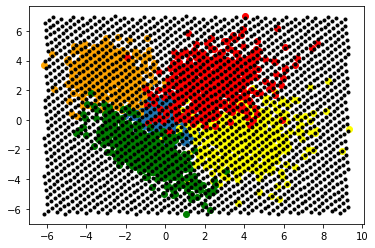}
  \caption{2500 sample points and 2048 potential representative points}
  \label{fig:p31}
\end{subfigure}
\hfill
\begin{subfigure}{0.3\textwidth}
    \includegraphics[width=\textwidth]{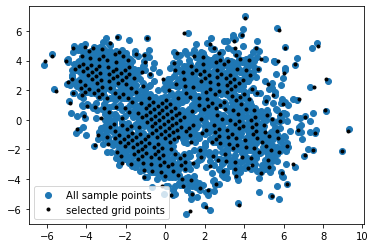}
  \caption{Mixed-Integer Solver (Gurobi with a Python extension) }
  \label{fig:p32}
\end{subfigure}
\hfill
\begin{subfigure}{0.3\textwidth}
    \includegraphics[width=\textwidth]{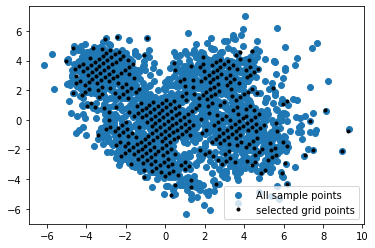}
  \caption{Subgradient method with momentum}
  \label{fig:p33}
\end{subfigure}
\hfill

\caption{$\text{dim}(\beta) = 5120000$, $\text{dim}(\gamma) = 2048$, and 409 selected particles}
\label{fig:P3}
\end{figure}

We provide results of experiments with the mixture Gaussian distribution, which imitates one step of the method \eqref{kernel-search}. This simple example, working with a 2-dimensional and 1-time-stage Gaussian distribution, demonstrates the advantage of the subgradient method over traditional state-of-the-art mixed-integer solvers such as Gurobi.
The marginal distribution $\tilde{\lambda}_t$ is supported
on five points $z^s$, and the conditional distributions $Q_t(\cdot|z^s)$, $s=1,\dots,5$, are normal with the parameters:
\begin{gather*}
\mu_1=\begin{bmatrix}
0 \\
0
\end{bmatrix}, \quad \mu_2=\begin{bmatrix}
4 \\
-1
\end{bmatrix}, \quad \mu_3=\begin{bmatrix}
-3 \\
3
\end{bmatrix},\quad \mu_4=\begin{bmatrix}
2.5\\
2.5
\end{bmatrix},\quad \mu_5=\begin{bmatrix}
-1 \\
-2
\end{bmatrix}.\\
\sigma_1=\begin{bmatrix}
0.5 & -0.2 \\
-0.2 & 0.5
\end{bmatrix},\sigma_2=\begin{bmatrix}
2 & 0 \\
0 & 2
\end{bmatrix}, \sigma_3=\begin{bmatrix}
1 & -0.1 \\
-0.1 & 1
\end{bmatrix}, \sigma_4=\begin{bmatrix}
2 & 0.5 \\
0.5 & 2
\end{bmatrix}, \sigma_5=\begin{bmatrix}
1.6 & -1.2 \\
-1.2 & 1.6
\end{bmatrix}.
\end{gather*}
We set $\alpha^{(0)} = 0.01$, $\epsilon = 10^{-7}$, $\varkappa_1 =0.35$, and $\varkappa_2 =0.35$. The potential representative points $\{\zeta^k\}_{k=1,\dots,K}$ were Sobol lattice points. For illustration, we use the lattice points that cover the entire graph, even if some are obviously not necessary. To find the optimal values of $\beta$ and $\gamma$ in problem \eqref{mixed-bin}, we used the mixed integer programming (MIP) solver Gurobi and Algorithm 1. In Figures \ref{fig:P1}--\ref{fig:P3}, the subfigures (a) show the sample points $\{x^{si}\}$ in five colors corresponding to the five Gaussian distributions and the potential locations of the representative particles. The subfigures (b) and (c) display the sample points and the grid points $\{z^k\}$ (black dots) selected
by the MIP solver and the subgradient method, respectively.  Table~\ref{gmm}  provides the total numbers of the variables $\beta$ and $\gamma$, the solution times of both methods (in seconds), and the values of the Wasserstein distance $W_1$ of the solutions obtained to the
colored cloud of particles. As the number of variables increases, the MIP solver takes an increasingly long time and becomes inapplicable. We further evaluated the effectiveness of the subgradient method on the multivariate Gaussian distribution and reported the results in Table \ref{mgd}, including the distribution's dimension. We also provide duality gap estimates, obtained as sketched below
\eqref{dual-problem}.

All numerical results were obtained using Python (Version 3.7) on a Macintosh HD laptop with a 2.9 GHz CPU and 16GB memory. In none of the experiments, the \emph{stochastic} subgradient method (sketched on p. 6) was competitive.

\begin{table}[H]

\caption{Comparison of the MIP solver and the subgradient method}

\begin{center}
\label{gmm}
\begin{tabular}{ cccccc}
\toprule
 dim($\beta$) &  dim($\gamma$)  & MIP (s) & subgradient (s) & MIP $W_1$ & subgradient  $W_1$\\
    \midrule
 128000 & 256  & 5.17 & 0.96 & 0.654 & 0.644\\
 512000 & 512  & 60.31 & 18.09 & 0.470 & 0.485\\
 5120000 & 2048  & 556.26 & 346.57 & 0.246 & 0.272\\
  20480000 & 4096  & - & 5130.23 & - & 0.222 \\
    \bottomrule
\end{tabular}
\end{center}

\end{table}

\begin{table}[H]

\caption{ Grid point selection with the subgradient method on multivariate Gaussian distribution }

\begin{center}
\label{mgd}
\begin{tabular}{ cccccc}
\toprule
 dim & dim($\beta$) &  dim($\gamma$)  & subgradient (s)  & subgradient  $W_1$ & duality gap\\
    \midrule
 3 & 20000000 & 4000  &  2661 & 0.361 & 0.01208\\
 4 & 31500000 & 4500  & 19493 & 0.564 & 0.00109\\
 5 & 40000000 & 5000  & 11922 & 0.830 & 0.00388\\
    \bottomrule
\end{tabular}
\end{center}

\end{table}

\begin{comment}

Figures \ref{fig:p1} and \ref{fig:p2} illustrate an example of selecting grid points from the two-dimensional multivariate normal distribution. In the grid selection method, we set the number of grid points to be around $M = 300$ selected out of 2500 randomly sampled points and 1600 potential grid points. We set $\alpha_0 = 0.01$, $\epsilon = 0.01$, $\varkappa_1 = 0.25$, and $\varkappa_2 = 0.001$. The gradient ascent algorithm with momentum converges in 1159 iterations.

\begin{figure}[h]
\centering

\begin{subfigure}{0.4\textwidth}
    \includegraphics[width=\textwidth]{figures/p1.png}
  \caption{2500 sample points and 1600 potential grid points}
  \label{fig:p1}
\end{subfigure}
\hfill
\begin{subfigure}{0.4\textwidth}
    \includegraphics[width=\textwidth]{figures/p2.png}
  \caption{352 selected grid points}
  \label{fig:p2}
\end{subfigure}
\hfill

\caption{}
\label{fig:p}
\end{figure}

\end{comment}

\section{Conclusion}
\label{conclusion}

The integrated transportation distance provides possibilities to approximate a large-scale Markov system with a simpler system and a finite state space. Based on this distance metric, we proposed a novel particle selection method that iteratively approximates the forward system stage-by-stage by utilizing our kernel distance. The heart of the method is a decomposable and parallelizable subgradient algorithm for particle selection, designed to circumvent the complexities of dealing with constraints and matrix computations.

To empirically validate our approach, we provide a straightforward example involving a 2-dimensional and 1-time-stage Gaussian distribution. We selected this simple case to aid in visualizing outcomes, enabling effective method comparisons and highlighting the limitations of Mixed Integer Programming (MIP) solvers in more complicated scenarios.

Additionally, it's worth noting that the integrated transportation distance and the particle selection method hold significant potential for various applications, especially with the aid of the dual subbgradient method. One such application pertains to look-ahead risk assessment in reinforcement learning, specifically in the context of Markov risk measures. Evaluating Markov risk measures in dynamic systems can be achieved through the equation \eqref{DP-risk-finite}, offering superior performance over one-step look-ahead methods. This approach streamlines risk or reward evaluation across a range of scenarios by substituting the approximate kernel in place of the original equation \eqref{DP-risk-finite}. We intend to explore the full spectrum of potential applications for this work in future research endeavors.

\bibliographystyle{abbrv}

%\bibliography{iclr2024_conference}

\newcommand{\noop}[1]{}

\end{document}